\documentclass{article}

\usepackage[preprint, nonatbib]{neurips_2023}

\usepackage[utf8]{inputenc} 
\usepackage[T1]{fontenc}    
\usepackage{hyperref}       
\usepackage{url}            
\usepackage{booktabs}       
\usepackage{amsfonts}       
\usepackage{nicefrac}       
\usepackage{microtype}      
\usepackage{xcolor}         
\usepackage{amsmath}

\usepackage{multicol}
\usepackage{multirow}
\usepackage{float}
\usepackage{makecell}
\usepackage{graphicx}
\usepackage{colortbl}
\usepackage{subcaption}
\usepackage{booktabs}

\title{\textit{Have Large Language Models Developed a Personality?}: Applicability of Self-Assessment Tests in Measuring Personality in LLMs}

\author{%
  Xiaoyang Song\\
  Data Science Institute\\
  Columbia University\\
  \texttt{xs2485@columbia.edu} \\
  \And
  Akshat Gupta\\
  J.P. Morgan AI Research\\
  \texttt{akshat.x.gupta@jpmorgan.com} \\
  \And
  Kiyan Mohebbizadeh\\
  Data Science Institute\\
  Columbia University\\
  \texttt{km3826@columbia.edu} \\
  \AND
  Shujie Hu \\
  Data Science Institute\\
  Columbia University\\
  \texttt{sh4355@columbia.edu} \\
  \And
  Anant Singh \\
  Department of Computer Engineering\\
  New York University\\
  \texttt{anant.singh@nyu.edu} \\
}

\begin{document}

\maketitle

\begin{abstract}
Have Large Language Models (LLMs) developed a personality? The short answer is a resounding "We Don't Know!". In this paper, we show that we do not yet have the right tools to measure personality in language models. Personality is an important characteristic that influences behavior. As LLMs emulate human-like intelligence and performance in various tasks, a natural question to ask is whether these models have developed a personality. Previous works have evaluated machine personality through self-assessment personality tests, which are a set of multiple-choice questions created to evaluate personality in humans. A fundamental assumption here is that human personality tests can accurately measure personality in machines. In this paper, we investigate the emergence of personality in five LLMs of difference sizes ranging from 1.5B to 30B. We propose the \textit{Option-Order Symmetry} property as a necessary condition for reliability of these self-assessment tests. Under this condition, the answer to self-assessment questions are invariant to the order in which the options are presented. We find that many LLMs personality test responses do not preserve option-order symmetry. We then take a deeper look at LLMs test responses where option-order symmetry is preserved to find that in these cases, LLMs do not take into account the situational statement being tested and produce the exact same answer irrespective of the situation being tested. We also identify the existence of inherent biases in these language models which is the root cause of the aforementioned phenomenon and makes self-assessment tests unreliable. These observations indicate that self-assessment tests are not the correct tools to measure personality in language models. Through this paper, we hope to draw attention to the shortcomings of current literature in measuring personality in LLMs and call for developing specific tools for machine personality measurement.

\end{abstract}

\section{Introduction}

Large Language Models (LLMs) have undergone significant advancements and demonstrated promising performance across various Natural Language Processing (NLP) tasks. These models are able to do traditional NLP tasks with ease without the need for any supervised training data \cite{gpt2}\cite{gpt3}\cite{gpt3.5}\cite{gpt4}. GPT-4, the most recent release by OpenAI, has reached human-level performance on academic and professional benchmarks \cite{gpt4}. A recent study has even referred to the near-human level capabilities of GPT-4 as "sparks of artificial general intelligence" \cite{sparks}. As these language models approach human-like language understanding capabilities and skills, the question arises: "Have these large language models also developed a personality?". This question might seem deceivingly simple at first, but it can have multiple interpretations. One interpretation of the above question is - do LLMs possess human-like qualities such as emotions, self-awareness, and consciousness? This interpretation is not the focus of our inquiry. Instead, we adopt another interpretation of the above question - have we created machines that display consistent patterns of behavior, and if so, can we quantify these behaviors? Under this interpretation, we are concerned with the behavioral patterns of LLMs as they manifest in the real world. In psychology, it is widely acknowledged that personality influences behavior \cite{personality_apa}. An individual's unique personality is a result of a very complex interplay between different biological factors like genetics or sex with other factors like life experiences, socio-cultural norms, etc. The underlying causes of machine personality, on the other hand, may differ significantly. In the case of large language models, it could be influenced by their specific architecture, training objectives, model size, and training dataset. In this paper, we do not delve into the underlying mechanisms that give rise to machine personality. Instead, our focus is primarily on how we can measure and quantify personality in language models.

As LLMs continue to improve in language understanding, and with the advent of ChatGPT, their integration into our daily lives becomes increasingly evident. These models, especially ChatGPT, are now frequently being used as brainstorming partners \cite{chatgpt}\cite{chatgpt_video}, and we interact with these systems constantly in natural language. These models serve as guides for decision-making, both in minor and significant aspects of our daily lives. With the emergence of agents like Auto-GPT \cite{auto-gpt} built on top of these LLMs, automated decisions will be made based on the outputs of such models. As these machines approach human-like language understanding capabilities \cite{gpt3.5}\cite{gpt4}, exhibiting sparks of intelligence and creativity \cite{sparks}, it becomes important to comprehend the nature of the entities we interact with and how they will behave in different situations. Studying personality in language models lies at the core of understanding their behavioral tendencies.

Personality is a significant characteristic that influences behavior and is characterised by personality traits \cite{personality_apa}\cite{personalitytraits_apa}. Personality traits can be conceptualized as distinct dimensions along which personality is categorized and measured. Personality traits in humans are measured using self-assessment personality tests. In the most popular version of these self-assessment tests, personality is evaluated across the Big Five factors of personality \cite{digman1990personality}\cite{goldberg1990alternative}\cite{goldberg1993structure}\cite{wiggins1996five}\cite{bigfive}, where a subject is tested across five dimensions or traits - openness, conscientiousness, extroversion, agreeableness, and neuroticism. These tests consist of various questions describing various situations corresponding to each personality trait in a multiple choice format following the Likert scale \cite{sullivan2013analyzing}\cite{human-symmetry}\cite{human-symmetry2}, which asks for the test taker's agreement to a question on usually a 5 or 7 point scale, ranging from strong agreement to strong disagreement. Table \ref{table:template} presents an example question template in which a situation is presented to the test taker, along with some example situations for each trait.

Previous research has utilized self-assessment tests designed for humans to measure personality in LLMs \cite{ai-personification}\cite{mpi}\cite{unc}. The underlying assumption is that \textit{the same tools used to evaluate human personality can be applied to assess machine personality}. In this paper, we question the applicability of these self-assessment tests for measuring personality in LLMs. We discover that LLM test responses to personality self-assessment questions lack a crucial property for test reliability - \textit{option-order symmetry}. While we do not rule out the possibility of presence of a consistent personality in LLMs, we find that it cannot be reliably measured using self-assessment tests created for humans. This highlights the necessity of developing improved tools to analyze and measure personality in language models. Specifically, our contributions in this paper are as follows:

\begin{itemize}
    \item We investigate the emergence of personality in LLMs as an outcome of the next-word prediction pre-training objective. We present personality self-assessment results for GPT2, GPT-Neo, and OPT models, ranging from a model size of 1.5B to 30B parameters. 
    \item We propose the property of \textit{option-order symmetry} as a necessary condition for reliability of personality self-assessment tests.
    \item We demonstrate that LLM self-assessment responses do not follow \textit{option-order symmetry}, rendering any conclusions drawn from these responses unreliable. We also present that LLMs' responses to self-assessment questions disregard the input situation being tested, underscoring the lack of effectiveness of these tests in measuring personality in LLMs.
    \item We identify the significant biases in these language models, which drive them to exhibit inherent preferences for certain choices over the others and thus cause the aforementioned issues. We present these biases by examining the model's output probability vector to the so-called \textit{content-free} templates and calibrate the language model output using them. 
\end{itemize}

\section{Related Work}\label{sec:relation-work}

\subsection{Human Personality Assessment}
\label{sec: bg}

Personality, as defined by the American Psychology Association \cite{personality_apa}, is "the enduring characteristic and behavior that comprise a person’s unique adjustment to life, including major traits, interests, drives, values, self-concept, abilities, and emotional patterns. Various theories explain the structure and development of personality in different ways, but all agree that personality helps determine behavior." Personality of an individual is evaluated across different personality traits. Personality traits are  stable, consistent, and consistent internal characteristics of an individual inferred from an individual's patterns of behaviors, attitudes, feelings, and habits \cite{personalitytraits_apa}. Personality traits can be thought of as principal components across which the personality of an individual is categorized and measured. Various theories in personality psychology have attempted to define the precise number of personality traits that can fully describe the personality of an individual. The most widely accepted taxonomy of personality traits is the \textit{Big Five} personality traits \cite{digman1990personality}\cite{goldberg1990alternative}\cite{goldberg1993structure}\cite{wiggins1996five}\cite{bigfive}, often referred to as OCEAN, which stands for Openness, Conscientiousness, Extroversion, Agreeableness, and Neuroticism. More detailed information about these personality traits can be found in the \ref{apdsec: personality-details}. 

The goal of personality assessments is to quantify one’s personality across the dimensions of predefined personality traits. One common method for conducting such assessments is through self-assessment tests. Self-assessments consist of a series of prompts that are associated with exactly one of the five traits. The subject is then required to choose a response that best reflects their tendencies on a Likert scale (typically a 5-point scale). For instance, consider a typical test question related to the trait of extroversion: the subject is presented with a situation - “I like being in crowded social situations”, and is asked to select from five ordinal options: very accurate (VA), moderately accurate (MA), neither accurate nor inaccurate (NANI), moderately inaccurate (MI), very inaccurate (VI). The selected option is mapped to a scale of 5 to 1 in this specific scenario. Once all the questions are answered,  statistical measures such as the mean and standard deviation of the OCEAN score for each trait are calculated for personality quantification.

\begin{table}[h]
    \begin{minipage}{.4\linewidth}
      \centering
    \begin{tabular}{l}
    \toprule
    \textbf{Example Template}\\
    \midrule
        Given a statement of you: "You \{\texttt{Situation}\}"\\
        Please choose from the following options to identify\\ how accurately this statement describes you. \\
        Options:\\
        (A). Very Accurate\\
        (B). Moderately Accurate\\
        (C). Neither Accurate Nor Inaccurate \\
        (D). Moderately Inaccurate\\
        (E). Very Inaccurate\\
        Answer: I choose option  \\
        \bottomrule
    \end{tabular}%
    \end{minipage}%
    \begin{minipage}{0.75\linewidth}
      \centering
        \begin{tabular}{l c c}
        
        \toprule
        \textbf{Example Situations} & \textbf{Trait} & \textbf{Key}\\
        \midrule
        Love to daydream & O & $+$\\
        Dislike changes & O & $-$\\
        Work hard & C & $+$\\
        Break rules & C & $-$\\
        Make friends easily & E & $+$\\
        Prefer to be alone & E & $-$\\
        Trust others & A & $+$\\
        Yell at people & A & $-$\\
        Worry about things & N & $+$\\
        Rarely overindulge & N & $-$\\
        \bottomrule
        \end{tabular}
    \end{minipage} 
    \vskip 0.1in
    \caption{The left table provides an example MPI template. A test question is formatted by replacing \{\texttt{Situation}\} with those situational statements introduced in the inventory \cite{mpi}. Each self-description is either positively or negatively related to the trait it measures as presented in the right table. }\label{table:template}
\end{table}

\subsection{Language Model Personality}
While language models have been used to predict the personality of human beings \cite{mehta2020bottom}\cite{christian2021text}, there have been very few studies about the development of personality in language models themselves. To the best of our knowledge, there are only three studies about language model personality. These are based in a setting where a language model takes the Big Five personality self-assessment tests and base their conclusions on the results of these tests. In this paper, we present evidence showing that these self-assessment tests are not the correct tool to measure personality in language models.

Karra et al. \cite{ai-personification} use a language inference (NLI) model trained over a BERT \cite{bert} model to detect language model personality. Here, the input questions from the Big Five test are used as the premise and each of the answer options is used as hypothesis, and the model classifies the hypothesis as being true (entailment) or false (contradiction) with respect to the given premise. The probability score corresponding to the entailment is then linearly interpolated between 1 to 5 to get a score on the Likert scale. This method has various limitations. Firstly, this method uses a secondary classifier (BERT \cite{bert}) as the NLI model, which brings in the biases of a secondary agent in the personality assessment results. Secondly, calibrating output probabilities of the NLI model to the Likert scale makes the results unreliable as there are no tests that confirm the existence of such calibration. Thirdly, this work does not test the (lack of) \textit{option-order symmetry} property (section \ref{sec:order-symmetry}), which checks the applicability of the self-assessment personality tests on language models. 

Jiang et al. \cite{mpi} and Caron et al. \cite{unc} overcome the limitation of using a secondary agent to evaluate personality in language models by having the LLMs answer these questions in a zero-shot setting identical to the one proposed in previous literature \cite{gpt2}\cite{gpt3}. While using this method does not corrupt the results of the personality self-assessment tests by an external agents, they still have their own limitations. Jiang et al.'s paper \cite{mpi} is the most comprehensive work so far on measuring personality in LLMs. We know that LLMs are very sensitive to templates and a template that is right for one model may not be right for another \cite{jiang2020can}\cite{cali}\cite{fantastically}. Also, the outputs corresponding to an \textit{optimal} template might be very different from an output corresponding to a \textit{bad} template. However, Jiang et al. \cite{mpi} do not sufficiently take these effects into account. Additionally, in their work \cite{mpi}, they use nucleus sampling \cite{nucleus-sampling} based generation to generate outputs for LLM personality responses. Due to sampling based generation, the answer to each of the questions given by an LLM is different every time the test is administered, resulting in very different results in each iteration. This makes the result of their tests unreliable, and cannot be replicated by a future study. We also find that the response generated by language models are not always perfectly interpretable, which thus makes this method not desirable and weakens the reliability of the personality tests. Examples of failure cases of using nucleus sampling-based generation on machine personality tests can be found in the \ref{apdsec: mcqa-intro-and-failure-case}. In our paper, we use the multiple-choice question-answering (MCQA) method inspired by previous literature \cite{gpt3}\cite{rae2021scaling}\cite{hoffmann2022training}\cite{llm-mcqa}, generating the most probable answer rather than sampling based generation. For this reason, we use open-sourced models for which we have full model access and can use beam-search to get the most probable answer instead of models with only blackbox access where we only get sampling based generation. 



\subsection{Multiple-Choice Question-Answering Using LLMs} \label{sec:mcqa}
Multiple-choice question-answering (MCQA) with LLMs is an important task on which a wide variety of LLMs \cite{gpt3}\cite{rae2021scaling}\cite{hoffmann2022training}\cite{gpt4} are regularly tested. Self-assessment tests follow an MCQA strategy called \textit{multiple choice prompting} (MCP) \cite{llm-mcqa}\cite{rae2021scaling}\cite{hoffmann2022training}. In this method, we prompt a model with a multiple-choice question with all the options provided as part of the question. Here, all the options are a part of the input prompt along with the question, which allows the model to project the answer to one of the available choices. The options are indexed by alpha-numeric symbols, and the answer is selected by selecting the index symbol with the highest probability \cite{rae2021scaling}\cite{hoffmann2022training}. 

A pre-requisite to extracting correct answers using this method is that the information about the option text is encoded in the option indexing symbol. This property is called the \textit{multiple-choice symbol-binding} (MCSB) property \cite{llm-mcqa}. The MCSB property is tested by shuffling the option texts in a question and checking how the answer to the questions are affected. For example, given the answer order “(A) Pizza, (B) Lollipop, (C) French beans” to a question, let's say that the highest probability is assigned to the token “A” which is associated with pizza. However, if we change the ordering of options to “(A) French beans, (B) Lollipops, (C) Pizza”, and the model still assigns the highest probability to “A,” which is now associated with French beans, we say that the model has low MCSB, which means that the option symbol is loosely bound to the option text. \cite{llm-mcqa} shows that even large LLMs like GPT-3 show low MCSB. This method of doing for MCQA with indexed templated is referred to as \textsc{MCP-Indexed} in the rest of paper. An example of \textsc{MCP-Indexed} template is shown in Table \ref{table:template}. 

To tackle the limitations posed when the option text is not bounded to the index symbol, we use the just option text to extract the answer. In this method, we still provide all the option choices in the input prompt, but extract the answer using option text. The option text for each option is appended to the input prompt (containing the question and all choices) one at a time, and the option corresponding to the highest probability is selected as the answer. To keep fair comparisons between the options of different lengths, the probability of each option is normalized by length. We follow \cite{gpt3} and take the $n^{th}$-root of the product of probabilities of option tokens to normalize the probabilities. The option indexing in this case is removed to eliminate any biases caused by the presence of indices. We call this method \textsc{MCP-Non-Indexed}. An example template would look like the template shown in Table \ref{table:template}, without the alphabet indexing.

\section{Datasets \& Models}\label{sec:dataset}
We conduct our experiment on MPI-1K dataset \cite{mpi}, which contains 1K self-assessment questions across the five \textsc{OCEAN} traits, created using the International Personality Item Pool \cite{ipip1}\cite{ipip2}\cite{ipip3}. The questions in the personality assessment test are of two types - positive and negative questions. The difference between these two types of questions is how they're scored. For a given trait, for example, \textit{openness}, if a question is positively correlated with the trait, for example - "You love to daydream", and the test taker selects "Very Accurate", it is scored as 5. Whereas if the question is negatively correlated with the trait - "You dislike changes" and the test taker selects "Very Accurate", it is scored as 1. Every trait has both positive and negative questions. Because of this, it is totally possible that the test-taking entity selects exactly one option blindly and still be able get a valid personality assessment score in terms of mean and standard deviation.

We present self-assessment results experiment with GPT2-XL-1.5B \cite{gpt2}, GPT-Neo-2.7B \cite{gpt-neo}, GPT-NeoX-20B \cite{gptneox} and OPT (13B \& 30B) \cite{opt}. These models are trained on huge amounts of natural language datasets and have demonstrated strong language understanding capabilities, making it reasonable to assume the existence of a consistent measurable personality. The choice of models also represents variations in pre-training datasets and model sizes. Overall, we examine 4 versions of GPT2 models, 4 versions of GPT-Neo models, and 8 versions of OPT models but only present the results of the mentioned 5 models in the main paper. Detailed results for each of these models can be found in the \ref{apdsec: model-details}. 

All pre-trained model checkpoints are obtained from Hugging Face Transformers library \cite{huggingface} and inferences are accelerated by one NVIDIA A100 80GB GPU for all models other than GPT-NeoX-20B and OPT-30B. For these large models, experiments are divided into mini-batches and carried out on 32 cores of Intel Xeon Platinum 8268 24C 205W 2.9GHz Processor in parallel.

\section{Experiments: LLM Personality Assessment}
In this section we go through the different steps involved in administering the self-assessment personality test for LLMs. Tests are administered using the MPI-1K dataset based on the Big Five personality test model \cite{goldberg1990alternative}\cite{bigfive}\cite{ipip-neo-120}. The aim of our experiments is to check the applicability of the human self-assessment tests for measuring personality in LLMs. The first step towards taking the personality test is to choose the right prompt template (section \ref{sec:prompt-selection}). We then present the property of \textit{option-order symmetry} as a fundamental property to be followed by a personality test response to be considered valid (section \ref{sec:order-symmetry}). Finally, we discuss that LLM self-assessment responses are unreliable measures of their personality and present that there exists internal biases in the language models which heavily influence the reliability of self-assessment tests (\ref{sec:calibration}). The code for reproducing these experimental results can be found in the supplementary materials.

\begin{table}
    \vskip 0.15in
    \centering 
    \scriptsize
    \setlength\tabcolsep{0pt}
    \setlength\extrarowheight{1pt}
    \begin{tabular*}{\textwidth}{@{\extracolsep{\fill\centering}}*{12}{c}}
        \toprule 
        \multirow{2}{*}[-0.3em]{\textsc{Model}} & 
            \multirow{2}{*}[-0.3em]{\textsc{Order}} &
            \multicolumn{2}{c}{\textbf{O}{\scriptsize penness}} & 
            \multicolumn{2}{c}{\textbf{C}{\scriptsize onscientiousness}} & 
            \multicolumn{2}{c}{\textbf{E}{\scriptsize xtraversion}} & 
            \multicolumn{2}{c}{\textbf{A}{\scriptsize greeableness}} & 
            \multicolumn{2}{c}{\textbf{N}{\scriptsize euroticism}} \\ 
        \addlinespace[0.125em] \cline{3-12} \addlinespace[0.25em]
        &  & Score & $\sigma$ &
            Score & $\sigma$ &
            Score & $\sigma$ &
            Score & $\sigma$ &
            Score & $\sigma$\\
        \midrule

        \multirow{2}{*}[-0em]{GPT2-XL-1.5B} & \textsc{Original}& $3.30$ & $1.63$ & $3.17$ & $1.87$ & $3.22$ & $1.85$ & $3.05$ & $1.68$ & $3.46$ & $1.77$ \\
        & \textsc{Reverse}& $2.95$ & $0.47$ & $2.88$ & $0.77$ & $3.06$ & $0.84$ & $3.07$ & $0.59$ & $2.97$ & $0.65$ \\
        \midrule
        \multirow{2}{*}[-0em]{GPT-Neo-2.7B} & \textsc{Original}& $3.29$ & $1.28$ & $2.98$ & $1.21$ & $3.18$ & $1.21$ & $3.10$ & $1.20$ & $2.75$ & $1.16$ \\
        & \textsc{Reverse}& $2.83$ & $1.86$ & $2.98$ & $1.92$ & $2.70$ & $1.83$ & $3.04$ & $1.88$ & $2.54$ & $1.82$ \\
        \midrule
        \multirow{2}{*}[-0em]{OPT-13B} & \textsc{Original}& $3.39$ & $1.92$ & $3.18$ & $1.92$ & $3.24$ & $1.87$ & $3.06$ & $1.83$ & $3.55$ & $1.86$ \\
        & \textsc{Reverse}& $2.62$ & $1.95$ & $2.91$ & $2.00$ & $2.77$ & $1.97$ & $3.08$ & $2.00$ & $2.38$ & $1.91$ \\
        \midrule
        \multirow{2}{*}[-0em]{GPT-NeoX-20B} & \textsc{Original}& $3.38$ & $1.97$ & $3.06$ & $1.99$ & $3.23$ & $1.98$ & $2.89$ & $1.99$ & $3.60$ & $1.91$ \\
        & \textsc{Reverse}& $2.62$ & $1.97$ & $2.90$ & $2.00$ & $2.72$ & $1.99$ & $3.08$ & $2.00$ & $2.38$ & $1.91$ \\
         \midrule
         \multirow{2}{*}[-0.0em]{OPT-30B} & \textsc{Original}& $2.83$ & $1.23$ & $3.14$ & $1.12$ & $2.88$ & $1.00$ & $3.06$ & $1.05$ & $2.67$ & $0.99$ \\
        & \textsc{Reverse}& $3.19$ & $0.98$ & $3.05$ & $1.00$ & $3.14$ & $0.99$ & $2.96$ & $1.00$ & $3.31$ & $0.95$ \\
        \midrule\midrule
        Human & - & $3.44$ & $1.13$ & $3.60$ & $0.98$ & $3.41$ & $1.07$ & $3.66$ & $1.04$ & $2.80$ & $1.06$\\
         \bottomrule
    \end{tabular*}
    \vskip 0.05in
        \caption{Selected models OCEAN scores and standard deviation with \textsc{MCP-Indexed} Template}\label{tbl:selected-score-indexed}
    \vskip -0.10in
\end{table}

\begin{table*}
    \vskip -0.1in
    \centering 
    \scriptsize
    \setlength\tabcolsep{0pt}
    \setlength\extrarowheight{1pt}
    \begin{tabular*}{\textwidth}{@{\extracolsep{\fill\centering}}*{11}{c}}
        \toprule 
        \multirow{2}{*}[-0.3em]{\textsc{Model}} & \multicolumn{5}{c}{Selected Options ($\%$)} & \multicolumn{5}{c}{OCEAN Scores Distribution}\\ 
        \addlinespace[0.125em] \cline{2-6} \cline{7-11} \addlinespace[0.25em]
        &  VA & MA & NANI & MI & VI & $5$ & $4$ & $3$ & $2$ & $1$\\
        \midrule 
        \multirow{1}{*}[-0em]{GPT2-XL-1.5B}  & $73.91$ & \cellcolor[HTML]{AAACED}$21.54$ & $4.55$ & $0.00$ & $0.00$ & $0.42$ & $0.11$ & $0.05$ & $0.11$ & $0.31$ \\
        \midrule

        \multirow{1}{*}[-0em]{GPT-Neo-2.7B}  & $16.89$ & $3.64$ & $0.00$ & \cellcolor[HTML]{AAACED}$79.47$ & $0.00$ & $0.11$ & $0.40$ & $0.00$ & $0.43$ & $0.06$ \\
        \midrule

        
        \multirow{1}{*}[-0em]{OPT-13B}  & \cellcolor[HTML]{AAACED}$87.46$ & $12.54$ & $0.00$ & $0.00$ & $0.00$ & $0.51$ & $0.05$ & $0.00$ & $0.08$ & $0.36$ \\
        \midrule
        
        \multirow{1}{*}[-0em]{GPT-NeoX-20B} & \cellcolor[HTML]{AAACED}$98.89$ & $0.00$ & $0.00$ & $1.11$ & $0.00$ & $0.55$ & $0.10$ & $0.00$ & $1.01$ & $0.44$ \\
        \midrule
        
        \multirow{1}{*}[-0em]{OPT-30B}  & $6.88$ & $0.00$ & $0.00$ & \cellcolor[HTML]{AAACED}$93.12$ & $0.00$ & $0.04$ & $0.41$ & $0.00$ & $0.52$ & $0.03$ \\
        \midrule\midrule
        Human & $14.80$ & $29.08$ & $18.98$ & $21.77$ & $15.37$ & $0.22$ & $0.32$ & $0.19$ & $0.18$ & $0.09$\\
         \bottomrule
    \end{tabular*}
        \caption{Selected models responses distribution and OCEAN score distribution with \textsc{MCP-Indexed} templates where options are presented in the \textsc{Original} order. For each model, the option that is colored in blue is the one that is preferred by its corresponding \textit{content-free} template defined in \ref{sec:calibration}.}\label{table:indexed-optionselection}
    \vskip -0.10in
\end{table*}


\begin{table*}[h]
    \vskip 0.05in
    \centering 
    \scriptsize
    \setlength\tabcolsep{0pt}
    \setlength\extrarowheight{1pt}
    \begin{tabular*}{\textwidth}{@{\extracolsep{\fill\centering}}*{12}{c}}
        \toprule 
        \multirow{2}{*}[-0.3em]{\textsc{Model}} & 
            \multirow{2}{*}[-0.3em]{\textsc{Order}} &
            \multicolumn{2}{c}{\textbf{O}{\scriptsize penness}} & 
            \multicolumn{2}{c}{\textbf{C}{\scriptsize onscientiousness}} & 
            \multicolumn{2}{c}{\textbf{E}{\scriptsize xtraversion}} & 
            \multicolumn{2}{c}{\textbf{A}{\scriptsize greeableness}} & 
            \multicolumn{2}{c}{\textbf{N}{\scriptsize euroticism}} \\ 
        \addlinespace[0.125em] \cline{3-12} \addlinespace[0.25em]
        &  & Score & $\sigma$ &
            Score & $\sigma$ &
            Score & $\sigma$ &
            Score & $\sigma$ &
            Score & $\sigma$\\
        \midrule 
        \multirow{2}{*}[-0em]{GPT2-XL-1.5B} & \textsc{Original} & $3.38$ & $1.97$ & $3.10$ & $2.00$ & $3.28$ & $1.99$ & $2.92$ & $2.00$ & $3.62$ & $1.91$ \\
        & \textsc{Reverse}& $3.38$ & $1.97$ & $3.10$ & $2.00$ & $3.28$ & $1.99$ & $2.92$ & $2.00$ & $3.62$ & $1.91$ \\
        \midrule
        
        \multirow{2}{*}[-0em]{GPT-Neo-2.7B} & \textsc{Original}& $3.38$ & $1.97$ & $3.10$ & $2.00$ & $3.28$ & $1.99$ & $2.92$ & $2.00$ & $3.62$ & $1.91$ \\
        & \textsc{Reverse}& $3.38$ & $1.97$ & $3.10$ & $2.00$ & $3.28$ & $1.99$ & $2.92$ & $2.00$ & $3.62$ & $1.91$ \\
        \midrule
        
        \multirow{2}{*}[-0em]{OPT-13B} & \textsc{Original}& $3.38$ & $1.97$ & $3.10$ & $2.00$ & $3.28$ & $1.99$ & $2.92$ & $2.00$ & $3.62$ & $1.91$ \\
        & \textsc{Reverse}& $3.38$ & $1.97$ & $3.10$ & $2.00$ & $3.28$ & $1.99$ & $2.92$ & $2.00$ & $3.62$ & $1.91$ \\
        \midrule
        \multirow{2}{*}[-0em]{GPT-NeoX-20B} & \textsc{Original}& $2.66$ & $1.98$ & $2.90$ & $2.00$ & $2.72$ & $1.99$ & $3.08$ & $2.00$ & $2.38$ & $1.91$ \\
        & \textsc{Reverse}& $2.62$ & $1.97$ & $2.90$ & $2.00$ & $2.72$ & $1.99$ & $3.08$ & $2.00$ & $2.38$ & $1.91$ \\
        \midrule
        \multirow{2}{*}[-0em]{OPT-30B} & \textsc{Original}& $3.38$ & $1.97$ & $3.10$ & $2.00$ & $3.28$ & $1.99$ & $2.92$ & $2.00$ & $3.62$ & $1.91$ \\
        & \textsc{Reverse}& $3.38$ & $1.97$ & $3.10$ & $2.00$ & $3.28$ & $1.99$ & $2.92$ & $2.00$ & $3.62$ & $1.91$ \\
        \midrule\midrule
        Human & - & $3.44$ & $1.13$ & $3.60$ & $0.98$ & $3.41$ & $1.07$ & $3.66$ & $1.04$ & $2.80$ & $1.06$\\
         \bottomrule
    \end{tabular*}
        \caption{Selected models OCEAN scores and standard deviation with \textsc{MCP-Non-Indexed} Template}\label{tbl:selected-score-nonindexed}
    \vskip -0.0in
\end{table*}

\begin{table*}[h]
    \centering 
    \scriptsize
    \setlength\tabcolsep{0pt}
    \setlength\extrarowheight{1pt}
    \begin{tabular*}{\textwidth}{@{\extracolsep{\fill\centering}}*{11}{c}}
        \toprule 
        \multirow{2}{*}[-0.3em]{\textsc{Model}} & \multicolumn{5}{c}{Selected Options ($\%$)} & \multicolumn{5}{c}{OCEAN Scores Distribution}\\ 
        \addlinespace[0.125em] \cline{2-6} \cline{7-11} \addlinespace[0.25em]
        &  VA & MA & NANI & MI & VI & $5$ & $4$ & $3$ & $2$ & $1$\\
        \midrule 
        \multirow{1}{*}[-0em]{GPT2-XL-1.5B} & \cellcolor[HTML]{AAACED}$100.00$ & $0.00$ & $0.00$ & $0.00$ & $0.00$ & $0.56$ & $0.00$ & $0.00$ & $0.00$ & $0.44$ \\
        \midrule

        \multirow{1}{*}[-0em]{GPT-Neo-2.7B}  & \cellcolor[HTML]{AAACED} $100.00$ & $0.00$ & $0.00$ & $0.00$ & $0.00$ & $0.56$ & $0.00$ & $0.00$ & $0.00$ & $0.44$ \\
        \midrule

        \multirow{1}{*}[-0em]{OPT-13B} & \cellcolor[HTML]{AAACED}$100.00$ & $0.00$ & $0.00$ & $0.00$ & $0.00$ & $0.56$ & $0.00$ & $0.00$ & $0.00$ & $0.44$\\
        \midrule

        \multirow{1}{*}[-0em]{GPT-NeoX-20B}  & $0.20$ & $0.00$ & $0.00$ & $0.00$ & \cellcolor[HTML]{AAACED}$99.80$ & $0.44$ & $0.00$ & $0.00$ & $0.00$ & $0.56$ \\
        \midrule

        \multirow{1}{*}[-0em]{OPT-30B} & \cellcolor[HTML]{AAACED}$100.00$ & $0.00$ & $0.00$ & $0.00$ & $0.00$ & $0.56$ & $0.00$ & $0.00$ & $0.00$ & $0.44$\\
        \midrule\midrule
        Human & $14.80$ & $29.08$ & $18.98$ & $21.77$ & $15.37$ & $0.22$ & $0.32$ & $0.19$ & $0.18$ & $0.09$\\
         \bottomrule
    \end{tabular*}
        \caption{Selected models responses distribution and OCEAN score distribution with \textsc{MCP-Non-Indexed} templates where options are presented in the \textsc{Original} order. For each model, the option colored in blue is the one that is preferred by its corresponding \textit{content-free} template defined in \ref{sec:calibration}}\label{table:nonindexed-optionselection}
    \vskip -0.10in
\end{table*}

\subsection{Unsupervised Prompt Selection}\label{sec:prompt-selection}
The first step in administering a self-assessment personality test for LLMs is to choose the right template to ask questions. One possible template is shown in Table \ref{table:template} for reference. Previous works \cite{jiang2020can}\cite{cali}\cite{fantastically} show that language model responses are sensitive to input prompts, and small differences in choices of words can lead to very different outputs. Optimal template selection is thus an important step. While significant amount work has been done on prompting, from learning optimal discrete prompts \cite{wallace2019universal}\cite{petroni2019language}\cite{jiang2020can}\cite{shin2020autoprompt} to learning continuous prompts \cite{li2021prefix}\cite{lester2021power}\cite{liu2021p} where additional parameters are learnt to augment language models, in our work we face a unique challenge. Each of the previous works use an annotated dataset to learn the correct discrete or continuous prompts. This means that selecting the right prompt depends on the availability of an annotated dataset of a collection of inputs and the corresponding expected outputs for those inputs. We call these methods \textit{supervised prompt selection} methods. In the scenario of self-assessment tests, there are no pre-existing expected outputs for a given input. Thus, prompt selection for self-assessment tests falls under the category of \textit{unsupervised prompt selection.}

To the best of our knowledge, only one paper in literature tries to tackle the problem of unsupervised prompt selection \cite{mi}. This procedure requires an unlabeled dataset of inputs and an output probability vector, where each dimension of the output vector is an output class. For example, if we have a sentiment classification task, the output vector would be two-dimensional, with the first dimension containing the probability of input corresponding to \textit{positive} sentiment while the second dimension corresponds to the probability of input corresponding to \textit{negative} sentiment. If we have a generation task, the number of output classes is the size of the vocabulary of the model. From a set of $N$ initial templates, this method chooses a prompt template that tries to maximize the \textit{mutual information} between the input data points and the output probability vector. This method was tested for various NLP tasks like classification, natural language inference, and question answering, and shows that high mutual information is the right metric to discover optimal templates in an unsupervised manner. For more information, we refer the reader to the original paper \cite{mi}.

We work with two types of templates - \textsc{MCP-Indexed} and \textsc{MCP-Non-Indexed} as described in section \ref{sec:mcqa}. An example of indexed templates is shown in Table \ref{table:template}. Here the options are indexed by alphabets from A to E. The non-indexed version of these templates would have no indexing available and the options would be listed with newline characters. For each indexing type experiment, we choose 36 initial templates following the principles introduced in the previous work \cite{mi}, and calculate mutual information values with a sample of 50 questions from the MPI-1K dataset, 10 questions from each trait. We select the template with the highest mutual information value for our experiments. The set of 36 templates was created based on the initial templates provided by Jiang et al. \cite{mpi}. More information about template creation rules, mutual information tests, and selected optimal templates for each model is presented in the \ref{apdsec: template-selection-intro-results}.

\subsection{Option-Order Symmetry}\label{sec:order-symmetry}
When we use personality self-assessment tests created for humans to study and quantify personality in LLMs, we make a fundamental assumption that these tests can measure personality in LLMs. For this to be true, an LLM test response should belong to the human response distribution and as a consequence follow the properties of this distribution. One such property of human test responses is the property of \textit{option-order symmetry}. This property states that the answer to a personality test question is invariant to the order in which the options are presented. Various studies \cite{human-symmetry}\cite{human-symmetry2} have shown that for human test takers, option-order symmetry is preserved between the original order of options (option order as shown in Table \ref{table:template}) and the reverse order of options (reversing the order of options in Table \ref{table:template}). As LLM personality is being interpreted via its human counterpart, we argue that this property should be preserved by LLM test responses for the response to be considered valid. This property is also an implicit test of the language understanding capability of language models. If option-order symmetry does not exist, a different ordering of options can result in different answers to the same situational question, thus questioning the reliability and trustworthiness of test results.

\textsc{\textbf{MCP-Indexed Templates :}} Table \ref{tbl:selected-score-indexed} shows personality assessment results for indexed templates for original and reverse order. 
Order symmetry is not found in any of the models that we experiment with. We consider these test responses as invalid and no conclusions can be made about the personality of LLMs based on these tests. Option-order symmetry also shows the presence of the property of MCSB for indexed templates.

Although the scores in Table \ref{tbl:selected-score-indexed} seem valid when compared to the average human scores shown in the last row, a look under the hood tells us that LLM test responses are unlike human test responses. The \textit{Selected Options} column in Table \ref{table:indexed-optionselection} shows the percentage of selected options for the different questions. For option-order-symmetric test responses, we see a high concentration of selection at one single option. For example, GPT-NeoX-20B chooses "Very Accurate" 98.89\% of time. It is still able to generate human-like mean scores as questions in personality tests are both positively and negatively correlated with the trait. Similarly, OPT-30B predominantly chooses "Moderately Inaccurate" for 93\% of the questions and is able to produce human-like standard deviations. This happens because of the presence of positively and negatively correlated questions in the self-assessment, which maps a single output response to two values as explained in section \ref{sec:dataset}. The corresponding output probability vectors for the scores are shown in the column \textit{OCEAN Scores Distribution}. This finding also invalidates the identifications of machine personality by matching its OCEAN scores to that of human, which is claimed in the previous work \cite{mpi}.


\textsc{\textbf{MCP-Non-Indexed Templates :}} Table \ref{tbl:selected-score-nonindexed} shows the personality assessment results when symbol indexing is removed. The test response distributions and score vectors are reported in Table \ref{table:nonindexed-optionselection}. We see that test responses of all models preserve option-order symmetry. We again see that all models select only one single option, this time with 100\% probability in 4 out of the 5 cases. We also see that 4 out of 5 models select "Very Accurate" as their chosen response. This selection answer option is independent of where the option "Very Accurate" appears in the prompt. For example, in the original-order prompt, the "Very Accurate" option occurs first, whereas, in the reverse-ordered prompt, this option occurs last. We also test these models with three other random orders of option presentation and find that "Very Accurate" is chosen every time. The results for all other models and the other random orders are presented in the \ref{apdsec: order-symmetry-results}. 

\begin{table*}
    \vskip 0.0in
    \centering 
    \scriptsize
    \setlength\tabcolsep{0pt}
    \setlength\extrarowheight{1pt}
    \begin{tabular*}{\textwidth}{@{\extracolsep{\fill\centering}}*{12}{c}}
        \toprule 
        \multirow{2}{*}[-0.3em]{\makecell[c]{\textsc{Model}\\(\textsc{Template})}} & 
            \multirow{2}{*}[-0.3em]{\textsc{Order}} &
            \multicolumn{2}{c}{\textbf{O}{\scriptsize penness}} & 
            \multicolumn{2}{c}{\textbf{C}{\scriptsize onscientiousness}} & 
            \multicolumn{2}{c}{\textbf{E}{\scriptsize xtraversion}} & 
            \multicolumn{2}{c}{\textbf{A}{\scriptsize greeableness}} & 
            \multicolumn{2}{c}{\textbf{N}{\scriptsize euroticism}} \\ 
        \addlinespace[0.125em] \cline{3-12} \addlinespace[0.25em]
        &  & Score & $\sigma$ &
            Score & $\sigma$ &
            Score & $\sigma$ &
            Score & $\sigma$ &
            Score & $\sigma$\\
        \midrule 
        \multirow{2}{*}[-0em]{GPT2-XL-1.5B} & \textsc{Original}& $3.19$ & $1.50$ & $3.42$ & $1.60$ & $3.32$ & $1.31$ & $3.13$ & $1.45$ & $3.02$ & $1.58$ \\
        & \textsc{Reverse}& $3.28$ & $1.75$ & $3.07$ & $1.80$ & $3.28$ & $1.70$ & $3.15$ & $1.77$ & $3.23$ & $1.79$ \\
        \midrule
        
        \multirow{2}{*}[-0em]{GPT-Neo-2.7B} & \textsc{Original}& $2.81$ & $1.92$ & $3.14$ & $1.92$ & $2.71$ & $1.94$ & $3.16$ & $1.96$ & $2.23$ & $1.81$ \\
        & \textsc{Reverse}& $2.96$ & $1.40$ & $3.60$ & $1.50$ & $3.07$ & $1.58$ & $3.51$ & $1.57$ & $2.77$ & $1.38$ \\

        \midrule
        \multirow{2}{*}[-0em]{OPT-13B} & \textsc{Original}& $2.97$ & $1.44$ & $2.97$ & $1.13$ & $3.02$ & $1.21$ & $2.77$ & $1.46$ & $3.21$ & $1.02$ \\
        & \textsc{Reverse}& $2.81$ & $1.07$ & $2.84$ & $1.14$ & $3.07$ & $0.77$ & $2.89$ & $1.08$ & $3.10$ & $0.86$ \\
        \midrule
        \multirow{2}{*}[-0em]{GPT-NeoX-20B} & \textsc{Original}& $3.28$ & $1.42$ & $3.41$ & $1.67$ & $3.09$ & $1.53$ & $3.25$ & $1.43$ & $2.95$ & $1.29$ \\
        & \textsc{Reverse}& $3.36$ & $1.39$ & $3.16$ & $1.64$ & $3.33$ & $1.43$ & $3.37$ & $1.48$ & $3.08$ & $1.43$ \\
        \midrule
        \multirow{2}{*}[-0em]{OPT-30B} & \textsc{Original}& $3.17$ & $1.51$ & $3.44$ & $1.34$ & $3.36$ & $1.40$ & $3.29$ & $1.59$ & $2.48$ & $1.68$ \\
        & \textsc{Reverse}& $3.19$ & $1.33$ & $3.15$ & $1.13$ & $3.01$ & $1.23$ & $3.14$ & $1.31$ & $2.56$ & $1.50$ \\
        \midrule\midrule
        Human & - & $3.44$ & $1.13$ & $3.60$ & $0.98$ & $3.41$ & $1.07$ & $3.66$ & $1.04$ & $2.80$ & $1.06$\\
         \bottomrule
    \end{tabular*}
        \caption{Selected models OCEAN scores and standard deviation with \textsc{MCP-Non-Indexed} Template after calibration using \textit{content-free} probabilities. Complete results can be found in the \ref{apdsec: calibration-results}.}\label{tbl:ocean-calibrated-non-indexed}
    \vskip -0.10in
\end{table*}

\subsection{Applicability of Self-Assessment Tests in Measuring Personality in LLMs}
\label{sec:calibration}
In previous section, we have seen two types of LLM self-assessment responses. In a few cases, LLM test responses do not possess option-order symmetry, thus rendering the results unreliable. In cases where option-order symmetry exists, we see that language models have a tendency to fixate on producing a single response for a large (if not complete) majority of input questions. Such a response does not depend on the choice of prompt template, the order in which the options are presented in the prompt, or the situational statement being tested in the prompt. It almost looks like LLMs are blindly selecting one single option irrespective of the situation being tested in the input, and have an inherent are biased towards that option. 

We test this hypothesis by passing a \textit{content-free} template to each of the models. To do this, we replace the situational statement in the template with an empty string and generate predictions for these content-free templates using same methods as before. The prediction for the content-free template are highlighted in blue in Tables \ref{table:indexed-optionselection} and \ref{table:nonindexed-optionselection}. We find that the predictions of the content-free templates are exactly what are predicted by the models even after the situational statement is inserted into the template. This shows that the model is (1) unable to take the semantic meaning of the inserted statement into account, and (2) that the model response is predetermined by the model template. This also indicates that it is highly likely that the symmetry that is observed on non-indexed template is due to the existence of these biases. We thus find that when LLMs are administered self-assessment tests, they either do not possess the option-order symmetry property, which makes the results of the test not trustworthy, or do not take into account the situational statements being tested, which defeats the purpose of these tests.

With this knowledge, we then re-calibrate the model outputs by scaling the each entry of output probabilities by its corresponding content-free probability. In this method, we leverage re-scaling to remove these initial biases towards certain choices which may be injected during the training stage. After calibration, we find that the models' choices are more diverse than before, and different options are being selected as what is observed in their human counterparts. Detailed results about response distributions can be found in the \ref{apdsec: calibration-results}. However, we find that symmetry is no longer preserved after calibration as shown in Table \ref{tbl:ocean-calibrated-non-indexed}., making these self-assessment tests unreliable. These results show that self-assessment personality tests created to measure human personality are not the correct tools to measure personality in LLMs.

\section{Conclusion}
As LLMs continue to mature and with their increasing incorporation in our daily lives, it is becoming more and more important to understand their behavioral tendencies. In this work, we present steps of administering self-assessment personality tests to measure personality in LLMs. We propose the property of \textit{option-order symmetry} as a necessary condition for the reliability of personality self-assessment tests for LLMs, which requires the answer to a self-assessment question to be invariant to the order in which the options are presented. We administer self-assessment personality tests for 5 LLMs of different sizes ranging from 1.5B to 30B parameters. We find that either option-order symmetry is not preserved, or the test responses are generated without taking into consideration the situation being tested. This indicates that self-assessment personality tests created for humans are not suitable to measure the personality of LLMs. Through these results, we call for a need to develop methods more suited to evaluate machine personality.

\section{Discussion}

Answering self-assessment questions is not a trivial task and requires many skills. The first skill is language understanding ability to interpret the question correctly. Next, the self-assessment question requires the subject to introspect and self-reflect. It requires the test taker to imagine themselves in a given situation, and understand how they would feel and react in that situation. The final step is to project their answers to one of the five available choices according to their subjective understanding of these scales. Can LLMs introspect and self-reflect to figure out how they would react in a given situation? Can LLMs figure out their own tendencies when inquired? Or is it better to present LLMs with a situation and ask them to react? 

LLMs are reactive agents. They are trained to react to given situations (input text) by producing text that follows the input text appropriately. We argue that to evaluate personality in LLMs, we should allow language models to do what they do best - react. An ideal personality test created for LLMs should present them with situations that do not require them to introspect to answer questions. Rather, it should just require the LLM to react to input text. An evaluation of LLM reactions to input situations would then be needed to analyze and quantify their behaviour. 

In addition to finding new ways of evaluating personality for machines, do we still keep evaluating machine personality using traits deemed suitable to describe human behavior? Perhaps the personality trait of extroversion is less important than the dominance trait for machines, which evaluates personality on the spectrum of being forceful to submissive. We not only need to rethink the personality evaluation method for machines, but we might also need to rethink the traits across which machine personality is evaluated. With this work, we want to implore researchers to start thinking about machine personality as an additional emergent outcome of the pre-training objective and how we can define and quantify it. 

\section*{Acknowledgements}
This paper was prepared for informational purposes in part by the Artificial Intelligence Research Group of JPMorgan Chase \& Co and its affiliates (“J.P. Morgan”) and is not a product of the Research Department of J.P. Morgan.  J.P. Morgan makes no representation and warranty whatsoever and disclaims all liability, for the completeness, accuracy, or reliability of the information contained herein.  This document is not intended as investment research or investment advice, or a recommendation, offer, or solicitation for the purchase or sale of any security, financial instrument, financial product, or service, or to be used in any way for evaluating the merits of participating in any transaction, and shall not constitute a solicitation under any jurisdiction or to any person if such solicitation under such jurisdiction or to such person would be unlawful. 

© 2022 JPMorgan Chase \& Co. All rights reserved.

\bibliographystyle{abbrv} 
\bibliography{llm_personality} 

\clearpage
\appendix

\section{Personality Assessment} 
\label{apdsec: personality-details}
\subsection{Personality Traits}

Lewis Goldberg created the Big 5 personality index \cite{bigfive} to measure, analyze, and compare human populations around the world. Goldberg first identified five traits (OCEAN traits) that comprise the human personality. Another contribution is that humans often do not lie at the extremes of personality scales, but somewhere in between. He created a test that made it possible to quantify these traits and determine a subject's unique personality \cite{Ackerman_2017}. Table \ref{apd:ocean-intro}. describes each OCEAN trait in detail.

\begin{table}[h]
    \centering 
    \small
    \begin{tabular}{l l}
    \toprule
    \hfil  OCEAN Trait & \hfil Description \\
        \midrule
        \textit{\textbf{O}penness - Closed Mindedness} & \makecell[l]{Willingness to embrace fresh ideas and novel experiences.}\\
        \midrule
        \textit{\textbf{C}onscientiousness - Lack of Direction} & \makecell[l]{Tendency to be organized, hard-working, goal-directed, and\\ adhere to social norms and rules.}\\
        \midrule
        \textit{\textbf{E}xtroversion - Introversion} & Measure of sociability, talkativeness, and excitability.\\
        \midrule
        \textit{\textbf{A}greeableness - Disagreeableness} & \makecell[l]{Measure of empathy, cooperativeness, politeness, kindness, and \\friendliness.}\\
        \midrule
        \textit{\textbf{N}euroticism - Emotional Stability} & Ability to balance and handle struggles in life.\\
        \bottomrule
    \end{tabular}
    \vskip 0.05in
    \caption{OCEAN trait introduction}\label{apd:ocean-intro}
    \vskip -0.1in
\end{table}

To measure these traits Goldberg expanded on common practice at the time of using scenario-based questions to test the subjects. Each prompt pertained to exactly one of the traits (positively or negatively). The subject then selected a choice describing how much they agreed with the prompt. An example of a prompt for positive extroversion would be "I feel comfortable attending parties where I do not know anyone". A negative version of this prompt would be "I do not feel comfortable at parties where I do not know anyone". These prompts are targeted towards exactly one of the traits (in this case extroversion) and represent behavior at the extremes of the spectrum for that trait. A prompt should not be associated with the middle of the trait spectrum, instead, the answer choices should then place the subject on the spectrum \cite{KWANTES2016229}.

The answer choices for these questions are based on a Likert scale. This scale should have an odd number of choices (so that there is a true neutral option) and typically should have a shorter rather than a larger one. A common practice for personality assessments is a 5-point scale with answer choices similar to \textit{Very Accurate, Moderately Accurate, Neither Accurate nor Inaccurate, Moderately Inaccurate, Very Inaccurate}. The answer choices should be evenly spaced about the neutral option. For example, there should not be three positive, one neutral, and one negative answer. In the case of personality assessment, the scaled values go from one to five with five being the positive extreme of the trait (a five in extroversion indicates that the subject is an extrovert). This also implies that for the negative prompts, the scale is reversed. Once the assessment has been conducted the mean of the scores for each trait is calculated resulting in five scores representing the subject's personality across the OCEAN traits \cite{McCrae_Costa_1987}.

These prompts have been carefully crafted by psychologists and are targeted toward exactly one of the five traits. As mentioned above the prompt is correlated positively or negatively to exactly one of the OCEAN traits \cite{KWANTES2016229}. Many of these prompts have been uploaded to the International Personality Item Pool (IPIP) which is an open-source database of prompts for various personality tests \cite{ipip_web}. From this database, researchers have collected a subset of 989 prompts to test machine personality tests. This subset includes the commonly used 50, 120, and 300 prompts from the original Big-5 test as well as the IPIP-NEO 120 and IPIP-NEO 300 tests \cite{mpi}. 

\subsection{Human Assessment}

Examiners must account for human behaviors and tendencies that could lead to inaccurate results. Human attention is a limited commodity. There must be measures taken to limit the interference that attention has on the results. This includes limiting the length of the assessment (number of questions) and following a logical structural protocol for the prompts and answers. This protocol includes consistent prompt structure, logical and consistent answer choice structure and clear instructions for the assessment \cite{human-symmetry, human-symmetry2, attention}. The other factor that must be considered is the bias toward societal expectations of each trait. Neuroticism tends to have a negative connotation while openness has a more positive connotation. This means that humans tend to answer prompts about neuroticism negatively while prompts about openness have a positive skew. Although this is more difficult to control in the assessment itself, it should be considered in the results.

\section{Models}
\label{apdsec: model-details}
\textbf{\textsc{OPT.}}
OPT \cite{opt} is a decoder-only pre-trained transformer with model sizes ranging from 125M to 175B parameters, proposed by Meta AI. It is a series of open-sourced models claimed to achieve comparable performance to GPT3 \cite{gpt3}. In our research, we experimented with OPT-125M, OPT-350M, OPT-1.3B, OPT-2.7B, OPT-6.7B, OPT-13B, and OPT-30B.

\textbf{\textsc{GPT2.}} GPT2 \cite{gpt2} is a unidirectional transformer pre-trained on a corpus of approximately 40 GB of text data. It is trained using a causal language modeling objective, which enhances its ability in  next token prediction. 
In our experiments, we explored models of all available sizes, including GPT2, GPT2-Medium, GPT2-Large, and GPT2-XL.

\textbf{\textsc{GPT-Neo.}} GPT-Neo \cite{gpt-neo} model is a GPT2-like language model trained on the Pile dataset that is introduced by the EleutherAI \cite{pile}. The architectural difference between GPT2 and GPT-Neo is that the latter uses local attention in every other layer with a window size of 256 tokens.
We included all sizes of GPT-Neo in our experiment, which are GPT-Neo-125M, GPT-Neo-1.3B, GPT-Neo-2.7B, as well as GPT-NeoX \cite{gptneox}, which is the largest 20B version.

\section{MCQA with LLMs} 
\label{apdsec: mcqa-intro-and-failure-case}
\subsection{Failure cases of conditional-generation-based MCQA}
In this work, we experiment with conditional-generation-based MCQA and find that it typically generates responses that are not interpretable.  This phenomenon is not restricted to the context of personality self-assessment tests but is generally true for MCQA tasks. Table \ref{apd:eg-failure}. provides two examples where conditional-generation-based MCQA with nucleus sampling fails. From these two examples, we observe that the generated responses, in this case, are either non-interpretable or contradictory. These failures are frequently observed in our experiments with different templates and different models.

Note that even though this phenomenon can be mitigated through prompt engineering and will not occur frequently as the model capacity grows, it still suffers from the indeterminacy involved in the decoding procedure. We will discard this MCQA method and instead use the most commonly used constraint MCQA strategy \cite{gpt3, llm-mcqa}. 
\begin{table}[h]
    \begin{minipage}{.5\linewidth}
      \centering
      \scriptsize
    \begin{tabular}{l}
    \toprule
    \small
    \textbf{Example 1:} Non-interpretable Responses\\
    \midrule
        Given a statement of you: "You are quiet around strangers." What \\
        do you think?\\
        Very Accurate\\
        Moderately Accurate\\
        Neither Accurate Nor Inaccurate \\
        Moderately Inaccurate\\
        Very Inaccurate\\
        Answer:  \\
        \midrule
        \textbf{\small Generated Response:} "You are quiet around strangers.\textbackslash n \\A: \textbackslash n
        You are quiet around strangers. This is a very common \\phrase. It is used to describe a person who is quiet around \\ people they don't know. It is also used to describe a person who\\ is quiet around people they do know."\\
        \bottomrule
    \end{tabular}%
    \end{minipage}%
    \begin{minipage}{0.5\linewidth}
      \centering
      \scriptsize
    \begin{tabular}{l}
    \toprule
    \small
    \textbf{Example 2:} Self-contradictory Responses\\
    \midrule
        Given a statement of you: "You worry about things." What \\
        do you think?\\
        Very Accurate\\
        Moderately Accurate\\
        Neither Accurate Nor Inaccurate \\
        Moderately Inaccurate\\
        Very Inaccurate\\
        Answer:  \\
        \midrule
        \textbf{\small Generated Response:} "You worry about things\textbackslash n\textbackslash n A:\textbackslash n\textbackslash n\\ You worry about things. \textbackslash n\textbackslash n is correct. \textbackslash n\textbackslash n You worry about \\things\textbackslash n\textbackslash n is incorrect."\\
        \bottomrule
    \end{tabular}%
    \end{minipage}
    \vskip 0.1in
    \caption{Two example failure cases of conditional-generation-based MCQA. Both results are obtained on GPT-Neo-2.7B using the nucleus sampling decoding strategy with the same set of parameters introduced by Jiang et al. \cite{mpi}: \texttt{top-p=0.95} and \texttt{temperaute=0.10}.}\label{apd:eg-failure}
\end{table}

\subsection{Constraint MCQA method \& Two template types}
We adopt the MCQA methods which are widely used in previous works \cite{gpt3, gpt3.5, llm-mcqa}, where we complete the question prompt with each choice and choose the one that gives the highest probability or the lowest perplexity score. We work with two types of templates: \textsc{MCP-Indexed} and \textsc{MCP-Non-Indexed}. Table \ref{apd:eg-mcp} shows examples of two templates. Specifically, for \textsc{MCP-Indexed} templates, we complete the question prompt with its formatted letter index of each choice, e.g. "(A).", while for \textsc{MCP-Non-Indexed} templates, we append the prompt with the description, e.g. "Very Accurate". In both cases, the score for each candidate option is calculated as the $n$th root of the product of the probabilities of appended option tokens. After all five options are examined, the output vector will be re-normalized to a probability vector and the option that associates with the highest probability is chosen as the answer.

\begin{table}[h]
    \begin{minipage}{.5\linewidth}
      \centering
      \scriptsize
    \begin{tabular}{l}
    \toprule
    \small
    \textbf{Example: \textsc{MCP-Indexed}} Template\\
    \midrule
        Given a statement of you: "You \{\texttt{Situation}\}." Please \\choose from the following options to identify how accurately\\this statement describes you. \\
        Options:\\
        (A). Very Accurate\\
        (B). Moderately Accurate\\
        (C). Neither Accurate Nor Inaccurate \\
        (D). Moderately Inaccurate\\
        (E). Very Inaccurate\\
        My answer: I choose option  \\
        \bottomrule
    \end{tabular}%
    \end{minipage}%
    \begin{minipage}{0.5\linewidth}
      \centering
      \scriptsize
    \begin{tabular}{l}
    \toprule
    \small
    \textbf{Example: \textsc{MCP-Non-Indexed}} Template\\
    \midrule
        Given a statement of you: "You \{\texttt{Situation}\}." Please \\choose from the following options to identify how accurately\\this statement describes you. \\
        Options:\\
        Very Accurate\\
        Moderately Accurate\\
        Neither Accurate Nor Inaccurate \\
        Moderately Inaccurate\\
        Very Inaccurate\\
        My answer: I choose option  \\
        \bottomrule
    \end{tabular}%
    \end{minipage}
    \vskip 0.1in
    \caption{Examples of two types of prompting templates.}\label{apd:eg-mcp}
\end{table}

\section{Unsupervised Template Selection}
\label{apdsec: template-selection-intro-results}
\subsection{Template Generation Principle}
We generate candidate templates based on the five different templates introduced by the previous work \cite{mpi}. We further decompose these templates into three parts as shown in Table \ref{apd:template}, where \textsc{Q-Prompt} and \textsc{A-Prompt} represent different prompt sentences used for questions and answers, respectively. Jiang et al. \cite{mpi} provided 3 types of \textsc{Q-Prompt} and 3 types of \textsc{A-Prompt} but examined only 5 possible combinations of them for template selection. In this work, we enumerate all possible combinations of these prompts and also introduce two more variations of the self-assessment test templates: (1) \textit{structured} templates and (2) \textit{lower-cased} templates.
\begin{table}[h]
    \centering 
    \small
    \begin{tabular}{l c}
    \toprule
    \hfil  Sample question of Machine Personality Inventory (MPI) & Components \\
        \midrule
        Given a statement of you: "\{\texttt{Situation}\}"
        Please choose from the following \\options to identify how accurately this statement describes you. & \textsc{Q-Prompt}\\
        Options:\\
        (A). Very Accurate\\
        (B). Moderately Accurate\\
        (C). Neither Accurate Nor Inaccurate & \textsc{Options}\\ 
        (D). Moderately Inaccurate\\
        (E). Very Inaccurate\\
        Answer: I choose option  & \textsc{A-Prompt}\\
        \bottomrule
    \end{tabular}
    \vskip 0.05in
    \caption{Introduction of each constituent in an example MPI-style question. Each question template contains three major components: \textsc{Q-Prompt}, \textsc{Options}, and \textsc{A-Prompt}, where each of them can be manipulated to form a new prompting template.}\label{apd:template}
    \vskip -0.1in
\end{table}

\textbf{\textsc{Structured Templates.}} Through introducing \textit{structured} template, we hope that adding spaces and structured words into the templates will help the model understand the self-assessment test question. Specifically, we add functional words, including \texttt{Question:}, \texttt{Options:}, and \texttt{Answer:} followed by extra empty newlines, before each component in the templates.

\textbf{\textsc{Lower-cased Templates.}} In \textit{lower-cased} template, we only modify the \textsc{Options} component by converting them to lower-case characters, by which we hope to reduce the noises and possibly misleading information introduced by those capital characters.

Combining together, we have 36 possible combinations of these variants, providing 36 candidates to perform template selection. Template details and notations are introduced in the next section.

\subsection{Template Notation}
Table \ref{apd:q-prompt}. and Table \ref{apd:a-prompt} shows the three types of \textsc{Q-Prompt} and three types of \textsc{A-Prompt} introduced by Jiang et al. \cite{mpi}, respectively.
\begin{table}[t]
    \centering 
    \small
    \begin{tabular}{l c}
    \toprule
    \hfil  \textsc{Q-Prompt} Templates & Type \& Notation \\

        \midrule
        Given a description of you: "You \{\texttt{Situation}\}." What do you think? & \textsc{Q-I}\\
        \midrule
        Given a description of you: "You \{\texttt{Situation}\}." What do you think?\\
        You can choose from the following options: & \multirow{1}{*}[0.3em]{\textsc{Q-II}}\\
        \midrule
        Given a statement of you: "You \{\texttt{Situation}\}."
        Please choose from the \\following options to identify how accurately this statement describes you. & \multirow{1}{*}[0.3em]{\textsc{Q-III}}\\
        \bottomrule
    \end{tabular}
    \vskip 0.05in
    \caption{Three types of \textsc{Q-Prompt} and their notations.}\label{apd:q-prompt}
    \vskip -0.1in
\end{table}
\begin{table}[!]
    \centering 
    \small
    \begin{tabular}{l c c}
    \toprule
    \hfil  \textsc{A-Prompt} Templates & Description & Type \& Notation \\

        \midrule
        My answer: & No guidance & \textsc{A-I}\\
        \midrule
        My answer: I choose option & Partial guidance & \multirow{1}{*}[0.0em]{\textsc{A-II}}\\
        \midrule
        My answer: I think the best description of myself is option & Full guidance &\multirow{1}{*}[0.0em]{\textsc{A-III}}\\
        \bottomrule
    \end{tabular}
    \vskip 0.05in
    \caption{Three types of \textsc{A-Prompt} and their notations. Note that in \textit{strutured} templates, \texttt{My answer:} will be replace by \texttt{Answer:} to make sure the overall prompting tone is consistent.}\label{apd:a-prompt}
    \vskip -0.1in
\end{table}In addition, \textit{structured} templates are denoted by "\textsc{s}" whereas the original non-structured templates will be abbreviated as "\textsc{ns}". Similarly, \textit{lower-cased} templates and original templates are denoted by "\textsc{lc}" and "\textsc{og}", respectively. With this notation, the template name is denoted by the concatenation of these labels. For instance, the \textit{structured} templates with \textsc{Q-I} type \textsc{Q-Prompt}, \textsc{A-II} type \textsc{A-Prompt}, and \textit{lower-cased} \textsc{Options} is represented by "\textsc{[lc]-[s]-[q-i]-[a-i]}". Table \ref{apd:template-example} provides four examples of these candidate templates.

\begin{table}[t]
    \begin{minipage}{.5\linewidth}
      \centering
      \scriptsize
    \begin{tabular}{l}
    \toprule
    \small
    \textbf{Template:} \textsc{[lc]-[ns]-[q-iii]-[a-ii]}\\
    \midrule
        Given a statement of you: "You \{\texttt{Situation}\}." Please \\choose from the following options to identify how accurately\\this statement describes you. \\
        Options:\\
        (A). very accurate\\
        (B). moderately accurate\\
        (C). neither accurate nor inaccurate \\
        (D). moderately inaccurate\\
        (E). very inaccurate\\
        My answer: I choose option  \\
        \bottomrule
    \end{tabular}%
    \end{minipage}%
    \begin{minipage}{0.5\linewidth}
      \centering
      \scriptsize
    \begin{tabular}{l}
    \toprule
    \small
    \textbf{Template:} \textsc{[og]-[ns]-[q-ii]-[a-iii]}\\
    \midrule
        Given a description of you: "You \{\texttt{Situation}\}." What do \\you think?\\
        You can choose from the following options:\\
        Options:\\
        (A). Very Accurate\\
        (B). Moderately Accurate\\
        (C). Neither Accurate Nor Inaccurate \\
        (D). Moderately Inaccurate\\
        (E). Very Inaccurate\\
        My answer: I think the best description of myself is option  \\
        \bottomrule
    \end{tabular}%
    \end{minipage}%
    \hfil
    \begin{minipage}{.5\linewidth}
    \vskip 0.10in
      \centering
      \scriptsize
    \begin{tabular}{l}
    \toprule
    \small
    \textbf{Template:} \textsc{[lc]-[s]-[q-iii]-[a-i]}\\
    \midrule
        Question:\\\\
        Given a statement of you: "You \{\texttt{Situation}\}." Please \\choose from the following options to identify how accurately\\this statement describes you. \\\\
        Options:\\
        (A). very accurate\\
        (B). moderately accurate\\
        (C). neither accurate nor inaccurate \\
        (D). moderately inaccurate\\
        (E). very inaccurate\\\\
        Answer:  \\
        \bottomrule
    \end{tabular}%
    \end{minipage}%
    \begin{minipage}{0.5\linewidth}
    \vskip 0.10in
      \centering
      \scriptsize
    \begin{tabular}{l}
    \toprule
    \small
    \textbf{Template:} \textsc{[og]-[s]-[q-ii]-[a-ii]}\\
    \midrule
        Question:\\
        \\
        Given a description of you: "You \{\texttt{Situation}\}." What do \\you think?\\
        You can choose from the following options: \\\\
        Options:\\
        (A). Very Accurate\\
        (B). Moderately Accurate\\
        (C). Neither Accurate Nor Inaccurate \\
        (D). Moderately Inaccurate\\
        (E). Very Inaccurate\\\\
        Answer: I choose option  \\
        \bottomrule
    \end{tabular}%
    \end{minipage}%
    \vskip 0.1in
    \caption{Four examples of generated candidate templates. The examples are presented under the \textsc{MCP-Indexed} type templates; for \textsc{MCP-Non-Indexed} type template, the generation principles remain unchanged but indices are dropped entirely.}\label{apd:template-example}
\end{table}

\subsection{Template Selection Criteria}
We perform template selection by calculating the \textit{Mutual Information} (MI) scores on a set of 50 questions drawn from the MPI-1K dataset, 10 questions for each OCEAN trait. Sorensen et al. \cite{mi} presents multiple-choice question-answering (MCQA) as a classification question, where the probability distribution $\mathbb{P}(Y\vert f(X))$ is of interest. In our context of self-assessment tests, $X$ are those \textit{situations} provided in the datasets and the five candidate options, $Y$ is the target choice, and $f(.)$ is the templatizing function which dictates how we prompt the question to the machine. Specifically, $f(.)$ can be thought of as a possible combination of prompts mentioned in the previous section, and $f(X)$ is the corresponding output probability vector of each option. With these notations, the mutual information score for a templatizing function $f_{i}(.)$ is denoted by $I(f_{i}(X); Y)$. \cite{mi}. As our work is a natural extension of previous work, we refer the readers to section 3.2 of the original paper \cite{mi} for more details about MI score approximation. 

For each model, we examine each possible templatizing function $f_i(.)$ (i.e. prompt template) and compute its MI scores, where the template with the highest score is selected as the \textit{optimal} template for the self-assessment test. The results of template selection are introduced in the next section.

\subsection{Template Selection Results}
In this section, we provide the template selection results for each model we experiment with. We perform template selection for both \textsc{MCP-Indexed} and \textsc{MCP-Non-Indexed} templates and results are reported in Table \ref{apd:template-selection-results}.

\begin{table}[H]
    \centering 
    \small
    \begin{tabular}{c c c}
    \toprule
    \textsc{Models} & \textsc{MCP-Indexed} & \textsc{MCP-Non-Indexed}\\
    \midrule
    OPT-125M & \textsc{[lc]-[s]-[q-i]-[a-i]} & \textsc{[lc]-[ns]-[q-ii]-[a-iii]}\\
    \midrule
    OPT-350M & \textsc{[lc]-[ns]-[q-iii]-[a-i]} & \textsc{[lc]-[ns]-[q-ii]-[a-ii]}\\
    \midrule
    OPT-1.3B & \textsc{[lc]-[ns]-[q-i]-[a-i]} & \textsc{[lc]-[s]-[q-ii]-[a-iii]}\\
    \midrule
    OPT-2.7B & \textsc{[og]-[s]-[q-iii]-[a-i]} & \textsc{[lc]-[s]-[q-ii]-[a-i]}\\
    \midrule
    OPT-6.7B & \textsc{[og]-[ns]-[q-iii]-[a-i]} & \textsc{[og]-[s]-[q-ii]-[a-iii]}\\
    \midrule
    OPT-13B & \textsc{[og]-[ns]-[q-ii]-[a-i]} & \textsc{[lc]-[s]-[q-iii]-[a-iii]}\\
    \midrule
    OPT-30B & \textsc{[lc]-[ns]-[q-i]-[a-i]} & \textsc{[lc]-[ns]-[q-ii]-[a-iii]}\\
    \midrule
    \midrule
    GPT2-Base-117M & \textsc{[og]-[s]-[q-ii]-[a-i]} & \textsc{[lc]-[ns]-[q-i]-[a-iii]}\\
    \midrule
    GPT2-Medium-345M & \textsc{[og]-[ns]-[q-iii]-[a-i]} & \textsc{[lc]-[ns]-[q-ii]-[a-iii]}\\
    \midrule
    GPT2-Large-774M & \textsc{[og]-[ns]-[q-iii]-[a-i]} & \textsc{[lc]-[s]-[q-i]-[a-iii]}\\
    \midrule
    GPT2-XL-1.5B & \textsc{[og]-[s]-[q-ii]-[a-i] } & \textsc{[lc]-[s]-[q-ii]-[a-ii]}\\
    \midrule
    \midrule
    GPT-Neo-125M & \textsc{[lc]-[s]-[q-i]-[a-i]} & \textsc{[og]-[ns]-[q-iii]-[a-i]}\\
    \midrule
    GPT-Neo-1.3B & \textsc{[og]-[ns]-[q-iii]-[a-i]} & \textsc{[lc]-[ns]-[q-ii]-[a-ii]}\\
    \midrule
    GPT-Neo-2.7B & \textsc{[lc]-[ns]-[q-i]-[a-iii]} & \textsc{[lc]-[s]-[q-ii]-[a-i]}\\
    \midrule
    GPT-NeoX-20B & \textsc{[lc]-[s]-[q-ii]-[a-i]} & \textsc{[og]-[s]-[q-ii]-[a-i]}\\
        \bottomrule
    \end{tabular}
    \vskip 0.05in
    \caption{\textit{Optimal} templates for both \textsc{MCP-Indexed} and \textsc{MCP-Non-Indexed} templates.}\label{apd:template-selection-results}
    \vskip -0.1in
\end{table}

\clearpage
\newpage
\section{\textsc{Option-Order Symmetry} Experimental Results}
\label{apdsec: order-symmetry-results}
\subsection{Choices of orders}
Robie et al. \cite{human-symmetry2} claim that option-order symmetry is expected in human personality tests, where human test-takers are expected to give similar responses even if the order of the provided options is changed. In experiments, we shuffle the orders of the options in the question prompt, and Table \ref{apd:orders}. provides details of the five orders that we experiment with. Note that for \textsc{MCP-Indexed} templates, only the descriptions are shuffled; in other words, the letter index remains in the order of A to E.

\begin{table}[H]
    \centering 
    \small
\begin{tabular}{c c}
    \toprule
    \textsc{Order} & \textsc{Example}\\
    \midrule
    \textsc{Original} & [VA, MA, NANI, MI, VI]\\
    \midrule
    \textsc{Reverse} & [VI, MI, NANI, MA, VA]\\
    \midrule
    \textsc{Order I} & [MA, VA, VI, MI, NANI]\\
    \midrule
    \textsc{Order II} & [VA, MA, NANI, MI, VI]\\
    \midrule
    \textsc{Order III} & [MA, VA, VI, MI, NANI]\\
        \bottomrule
    \end{tabular}
    \vskip 0.05in
    \caption{Five different option orders\label{apd:orders}}
    \vskip -0.1in
\end{table}

\newpage
\subsection{Complete Experimental Results}

\subsubsection{OPT models with \textsc{MCP-Indexed} Template: OCEAN scores}
\begin{table*}[h]
    \vskip 0.05in
    \centering 
    \scriptsize
    \setlength\tabcolsep{0pt}
    \setlength\extrarowheight{1pt}
        \caption{OPT models OCEAN scores and standard deviation with \textsc{MCP-Indexed} Template}\label{apd:complete-results-opt-indexed}
    \vskip -0.0in
\end{table*}
\clearpage
\subsubsection{OPT models with \textsc{MCP-Indexed} Template: Response distribution}
\begin{table*}[h]
    \vskip 0.15in
    \centering 
    \scriptsize
    \setlength\tabcolsep{0pt}
    \setlength\extrarowheight{1pt}
        \caption{OPT models responses and OCEAN score distribution with \textsc{MCP-Indexed} templates}
    \vskip -0.1in
\end{table*}
\clearpage
\subsubsection{GPT2 \& GPT-Neo models with \textsc{MCP-Indexed} Template: OCEAN scores}
\begin{table*}[h]
    \vskip 0.05in
    \centering 
    \scriptsize
    \setlength\tabcolsep{0pt}
    \setlength\extrarowheight{1pt}
        \caption{GPT2 \& GPT-Neo models OCEAN scores and standard deviation with \textsc{MCP-Indexed} Template}\label{apd:complete-results-gpt-indexed}
    \vskip -0.0in
\end{table*}
\clearpage
\subsubsection{GPT2 \& GPT-Neo models with \textsc{MCP-Indexed} Template: Response distribution}
\begin{table*}[h]
    \vskip 0.15in
    \centering 
    \scriptsize
    \setlength\tabcolsep{0pt}
    \setlength\extrarowheight{1pt}
        \caption{GPT2 \& GPT-Neo models responses and OCEAN score distribution with \textsc{MCP-Indexed} templates}
    \vskip -0.1in
\end{table*}

\clearpage
\subsubsection{OPT models with \textsc{MCP-Non-Indexed} Template: OCEAN scores}
\begin{table*}[h]
    \vskip 0.05in
    \centering 
    \scriptsize
    \setlength\tabcolsep{0pt}
    \setlength\extrarowheight{1pt}
        \caption{OPT models OCEAN scores and standard deviation with \textsc{MCP-Non-Indexed} Template}\label{apd:complete-results-opt-non-indexed}
    \vskip -0.0in
\end{table*}

\clearpage
\subsubsection{OPT models with \textsc{MCP-Non-Indexed} Template: Response distribution}
\begin{table*}[h]
    \vskip 0.15in
    \centering 
    \scriptsize
    \setlength\tabcolsep{0pt}
    \setlength\extrarowheight{1pt}
        \caption{OPT models responses and OCEAN score distribution with \textsc{MCP-Non-Indexed} templates}
    \vskip -0.1in
\end{table*}

\clearpage
\subsubsection{GPT2 \& GPT-Neo models with \textsc{MCP-Non-Indexed} Template: OCEAN scores}
\begin{table*}[h]
    \vskip 0.05in
    \centering 
    \scriptsize
    \setlength\tabcolsep{0pt}
    \setlength\extrarowheight{1pt}
        \caption{GPT2 \& GPT-Neo models OCEAN scores and standard deviation with \textsc{MCP-Non-Indexed} Template}\label{apd:complete-results-gpt-non-indexed}
    \vskip -0.0in
\end{table*}

\clearpage
\subsubsection{GPT2 \& GPT-Neo models with \textsc{MCP-Non-Indexed} Template: Response distribution}
\begin{table*}[h]
    \vskip 0.15in
    \centering 
    \scriptsize
    \setlength\tabcolsep{0pt}
    \setlength\extrarowheight{1pt}
        \caption{GPT models responses and OCEAN score distribution with \textsc{MCP-Non-Indexed} templates}
    \vskip -0.1in
\end{table*}

\clearpage
\section{\textsc{Calibration} Experimental Results} 
\label{apdsec: calibration-results}
\subsection{OPT models with \textsc{MCP-Non-Indexed} Template: OCEAN scores}
\begin{table*}[h]
    \vskip 0.05in
    \centering 
    \scriptsize
    \setlength\tabcolsep{0pt}
    \setlength\extrarowheight{1pt}
        \caption{OPT models OCEAN scores and standard deviation with \textsc{MCP-Non-Indexed} Template after calibration with \textit{content-free} probability}\label{apd:complete-results-opt-non-indexed-cali}
    \vskip -0.0in
\end{table*}

\clearpage
\subsection{OPT models with \textsc{MCP-Non-Indexed} Template: Response distribution}
\begin{table*}[h]
    \vskip 0.15in
    \centering 
    \scriptsize
    \setlength\tabcolsep{0pt}
    \setlength\extrarowheight{1pt}
        \caption{OPT models responses and OCEAN score distribution with \textsc{MCP-Non-Indexed} templates after calibration with \textit{content-free} probability}
    \vskip -0.1in
\end{table*}

\clearpage
\subsection{GPT2 \& GPT-Neo models with \textsc{MCP-Non-Indexed} Template: OCEAN scores}
\begin{table*}[h]
    \vskip 0.05in
    \centering 
    \scriptsize
    \setlength\tabcolsep{0pt}
    \setlength\extrarowheight{1pt}
        \caption{GPT2 \& GPT-Neo models OCEAN scores and standard deviation with \textsc{MCP-Non-Indexed} Template after calibration with \textit{content-free} probability}\label{apd:complete-results-gpt-non-indexed-cali}
    \vskip -0.0in
\end{table*}

\clearpage
\subsection{GPT2 \& GPT-Neo models with \textsc{MCP-Non-Indexed} Template: Response distribution}
\begin{table*}[h]
    \vskip 0.15in
    \centering 
    \scriptsize
    \setlength\tabcolsep{0pt}
    \setlength\extrarowheight{1pt}
        \caption{GPT2 \& GPT-Neo models responses and OCEAN score distribution with \textsc{MCP-Non-Indexed} templates after calibration with \textit{content-free} probability.}
    \vskip -0.1in
\end{table*}

\end{document}